\documentclass[11pt,a4paper]{article}
\usepackage[hyperref]{emnlp2020}
\usepackage{times}
\usepackage{latexsym}

\usepackage[disable]{todonotes}
\usepackage{multicol}
\usepackage{multirow}
\usepackage{booktabs}
\usepackage{natbib}
\usepackage{amsthm}
\usepackage{amssymb}
\usepackage{amsmath}
\usepackage{amsfonts}
\usepackage{xcolor}
\usepackage{microtype}
\usepackage{comment}
\usepackage{caption,subcaption}
\usepackage{tabularx}
\usepackage{colortbl}

\aclfinalcopy 

\definecolor{Gray}{gray}{0.85}
\definecolor{LightCyan}{rgb}{0.88,1,1}

\newcolumntype{a}{>{\columncolor{Gray}}c}
\title{\textit{From Zero to Hero}: On the Limitations of Zero-Shot Cross-Lingual Transfer with Multilingual Transformers}

\author{Anne Lauscher\textsuperscript{1}\thanks{~~Equal contribution.}, Vinit Ravishankar\textsuperscript{2}\footnotemark[1], Ivan Vuli\'{c}\textsuperscript{3}, and Goran Glava\v{s}\textsuperscript{1} \vspace{0.3em} \\
  \textsuperscript{1}Data and Web Science Group, University of Mannheim, Germany \\
  \textsuperscript{2}Language Technology Group, University of Oslo, Norway\\
  \textsuperscript{3}Language Technology Lab, University of Cambridge, UK \hspace{2mm} \vspace{0.3em} \\ 
  \textsuperscript{1}{\tt \{anne,goran\}@informatik.uni-mannheim.de}, \\
  \textsuperscript{2}{\tt vinitr@ifi.uio.no}, 
  \textsuperscript{3}{\tt iv250@cam.ac.uk}
}

\date{}

\begin{document}
\maketitle
\begin{abstract}
Massively multilingual transformers pretrained with language modeling objectives (e.g., mBERT, XLM-R) have become a \textit{de facto} default transfer paradigm for zero-shot cross-lingual transfer in NLP, offering unmatched transfer performance. Current downstream evaluations, however, verify their efficacy predominantly in transfer settings involving languages with sufficient amounts of pretraining data, and with lexically and typologically close languages. In this work, we analyze their limitations and show that cross-lingual transfer via massively multilingual transformers, much like transfer via cross-lingual word embeddings, is substantially less effective in resource-lean scenarios and for distant languages. Our experiments, encompassing three lower-level tasks (POS tagging, dependency parsing, NER), as well as two high-level semantic tasks (NLI, QA), empirically correlate transfer performance with linguistic similarity between the source and target languages, but also with the size of pretraining corpora of target languages. We also demonstrate a surprising effectiveness of inexpensive few-shot transfer (i.e., fine-tuning on a few target-language instances after fine-tuning in the source) across the board. This suggests that additional research efforts should be invested to reach beyond the limiting zero-shot conditions.
%


\end{abstract}

\section{Introduction and Motivation}
\label{s:intro}
Labeled datasets of sufficient size support supervised learning and development in NLP. However, given the notorious tediousness, subjectivity, and cost of linguistic annotation \cite{dandapat2009complex,sabou2012crowdsourcing,fort2016collaborative}, as well as a large number of structurally different NLP tasks, such data typically exist only for English and a handful of resource-rich languages \cite{bender2011achieving,Ponti:2019cl,joshi2020state}. The data scarcity issue renders the need for effective \textit{cross-lingual transfer} strategies: how can one exploit abundant labeled data from resource-rich languages to make predictions in resource-lean languages? 

In the most extreme scenario, termed \textit{zero-shot cross-lingual transfer}, not even a single annotated example is available for the target language. Recent work has placed much emphasis on the zero-shot scenario exactly; in theory, it offers the widest portability across the world's (more than) 7,000 languages \cite{pires_how_2019,artetxe2019cross,Lin:2019acl,cao2020multilingual,hu2020xtreme}.

The current mainstay of cross-lingual transfer in NLP are approaches based on continuous cross-lingual representation spaces such as cross-lingual word embeddings \cite{Ruder:2019jair} and, most recently, massively multilingual transformer models that are pretrained on multilingual corpora using language modeling objectives
\cite{devlin2019bert,lample2019cross,Conneau:2020acl}. The latter have \textit{de facto} become the default paradigm for cross-lingual transfer, with a number of studies reporting their unparalleled cross-lingual transfer performance \cite[\textit{inter alia}]{pires_how_2019,wu-dredze-2019-beto,ronnqvist2019multilingual,karthikeyan2020cross,wu2020emerging}.

\vspace{1.6mm}
\noindent \textbf{Key Questions and Contributions.} In this work, we dissect the current state-of-the-art approaches to (zero-shot) cross-lingual transfer, and analyze a variety of conditions and underlying factors that critically impact or limit the ability to conduct effective cross-lingual transfer via massively multilingual transformer models. We aim to provide answers to the following crucial questions. 

\vspace{1.3mm}
\noindent \textbf{(Q1)} \textit{What is the role of language (dis)similarity and language-specific corpora size for pretraining?}

\vspace{1.3mm}
\noindent Current cross-lingual transfer with massively multilingual models is still primarily focused on transfer to either (1) languages that are typologically or etymologically close to English (e.g., German, Scandinavian languages, French, Spanish), or (2) languages with large monolingual corpora, well-represented in the massively multilingual pretraining corpora (e.g., Arabic, Hindi, Chinese). Furthermore, \newcite{wu2020emerging} suggest that pretrained transformers, much like static word embedding spaces, yield language representations that are easily (linearly) alignable between languages, but limit their study to major languages: Chinese, Russian, and French. However, language transfer with static cross-lingual word embeddings has been shown ineffective when involving dissimilar languages \cite{sogaard2018limitations,vulic2019we} or languages with small corpora \cite{vulic2020all}.

We therefore probe pretrained multilingual models in diverse transfer settings encompassing more distant languages, and languages with varying size of pretraining corpora. We demonstrate that, similar to prior research in cross-lingual word embeddings, transfer performance crucially depends on two factors: (1) linguistic (dis)similarity between the source and target language and (2) size of the pretraining corpus of the target language.

\vspace{1.3mm}
\noindent \textbf{(Q2)} \textit{What is the role of a particular task in consideration for transfer performance?}

\vspace{1.3mm}
\noindent We conduct all analyses across five different tasks, which we roughly divide into two groups: (1) ``lower-level'' tasks (POS-tagging, dependency parsing, and NER); and (2) ``higher-level'' language understanding tasks (NLI and QA). We show that transfer performance in both zero-shot and few-shot scenarios largely depends on the ``task level''.

\vspace{1.3mm}
\noindent \textbf{(Q3)} \textit{Can we even predict transfer performance?}

\vspace{1.3mm}
\noindent 
Running a simple regression model on available transfer results, we show that transfer performance can (roughly) be predicted from the two crucial factors: linguistic (dis)similarity \cite{littell-etal-2017-uriel} is a strong predictor of transfer performance in lower-level tasks; for higher-level tasks such as NLI and QA, both factors seem to contribute.

\vspace{1.3mm}
\noindent \textbf{(Q4)} \textit{Should we focus more on few-shot transfer scenarios and quick annotation cycles?}

\vspace{1.3mm}
\noindent 
Complementary to the efforts on improving zero-shot transfer \cite{cao2020multilingual}, we point to few-shot transfer as a very effective mechanism for improving language transfer performance in downstream tasks. Similar to the seminal ``pre-neural'' work of \newcite{Garrette:2013naacl}, our results suggest that only several hours or even minutes of annotation work can ``buy'' a lot of performance points in the low-resource target tasks. For all five tasks in our study, we obtain substantial (and in some cases surprisingly large) improvements with minimal annotation effort. For instance, for dependency parsing, in some target languages we improve up to 40 UAS points by additional fine-tuning on as few as 10 target language sentences. Moreover, the few-shot gains are most prominent exactly where zero-shot transfer fails: for distant target languages with small monolingual corpora.

\section{Background and Related Work}
For completeness, we now provide a brief overview of \textbf{1)} different approaches to cross-lingual transfer, with a focus on \textbf{2)} the state-of-the-art massively multilingual transformer (MMT) models, and then \textbf{3)} position our work with respect to other studies that examine different properties of MMTs. 

\subsection{Cross-Lingual Transfer Methods in NLP}
In order to enable language transfer, we must represent the texts from both the source and target language in a shared cross-lingual representation space, which can be discrete or continuous. Language transfer paradigms based on discrete representations include \textit{machine translation} of target language text to the source language (or vice-versa) \cite{mayhew2017cheap,eger2018cross}, and grounding texts from both languages in \textit{multilingual knowledge bases} (KBs) \cite{navigli2012babelnet,lehmann2015dbpedia}. While reliable machine translation hinges on the availability of sufficiently large and in-domain parallel corpora, a prerequisite still unsatisfied for the vast majority of language pairs, transfer through multilingual KBs \cite{Camacho:2016nasari,Mrksic:2017tacl} is impaired by the limited KB coverage and inaccurate entity linking \cite{mendes2011dbpedia,moro2014entity,raiman2018deeptype}.  

Therefore, recent years have seen a surge of cross-lingual transfer approaches based on continuous cross-lingual representation spaces. The previous state-of-the-art approaches, predominantly based on cross-lingual word embeddings \cite{mikolov2013exploiting,Ammar:2016tacl,artetxe2017learning,smith2017offline,glavavs2019properly,vulic2019we} and sentence embeddings \cite{Artetxe:2019tacl}, are now getting replaced by massively multilingual transformers based on language modeling objectives \cite{devlin2019bert,lample2019cross,Conneau:2020acl}.

\subsection{Massively Multilingual Transformers}

\noindent \textbf{Multilingual BERT (mBERT).} At the core of BERT \cite{devlin2019bert} is a multi-layer transformer network \citep{vaswani2017attention}. Its parameters are pretrained using two objectives: masked language modeling (MLM) and next sentence prediction (NSP). In MLM, some percentage of tokens are masked out and they need to be recovered from the context. NSP predicts if two given sentences are adjacent in text and serves to model longer-distance dependencies spanning across sentence boundaries. \newcite{liu_roberta_2019} introduce RoBERTa, a robust variant of BERT pretrained on larger corpora, showing that NSP can be omitted if the transformer's parameters are trained with MLM on sufficiently large corpora. Multilingual BERT (mBERT) is a BERT model trained on concatenated multilingual Wikipedia corpora of 104 languages with largest Wikipedias.\footnote{\url{https://github.com/google-research/bert/blob/master/multilingual.md}} To alleviate underfitting (for languages with smaller Wikipedias) and overfitting, up-sampling and down-sampling are done via exponentially smoothed weighting.         



\vspace{1.6mm}
\noindent \textbf{XLM on RoBERTa (XLM-R).} XLM-R \cite{Conneau:2020acl} is a robustly trained RoBERTa, exposed to a much larger multilingual corpus than mBERT. It is trained on the CommonCrawl-100 data \cite{Wenzek:2019arxiv} of 100 languages. There are 88 languages in the intersection of XLM-R's and mBERT's corpora; for some languages (e.g., Kiswahili), XLM-R's monolingual data are several orders of magnitude larger than with mBERT.

\vspace{1.6mm}
\noindent \textbf{The ``Curse of Multilinguality''.} 
\newcite{Conneau:2020acl} observe the following phenomenon working with XLM-R: for a fixed model capacity, the cross-lingual transfer performance improves when adding more pretraining languages only up to a certain point. After that, adding more languages to pretraining degrades transfer performance. This effect, termed the ``curse of multilinguality'', can be mitigated by increasing the model capacity \cite{artetxe2019cross}. However, this also suggests that the model capacity is a critical limitation to zero-shot cross-lingual transfer, especially when dealing with lower computational budgets.

%

In this work (see \S\ref{sec:fewshot}), we suggest that a light-weight strategy to mitigate this effect/curse for improved transfer performance is abandoning the zero-shot paradigm in favor of few-shot transfer. If one targets improvements in a particular target language, it is possible to obtain large gains across different tasks at a very small annotation cost in the target language, and without the need to train a larger-capacity MMT from scratch.  




\subsection{Cross-Lingual Transfer with MMTs}

A number of recent BERTology efforts,\footnote{A name for the body of work analyzing the abilities of BERT and the knowledge encoded in its parameters.} which all emerged within the last year, aim at extending our understanding of the knowledge encoded and abilities of MMTs. \newcite{libovicky2020language} analyze language-specific versus language-universal knowledge in mBERT. \newcite{pires_how_2019} show that zero-shot cross-lingual transfer with mBERT is effective for POS tagging and NER, and that it is more effective between related languages. \newcite{wu-dredze-2019-beto} extend the analysis to more tasks and languages; they show that transfer via mBERT is competitive to the best task-specific zero-shot transfer approach in each task. Similarly, \citet{karthikeyan2020cross} prove mBERT to be effective for NER and NLI transfer to Hindi, Spanish, and Russian (note that all languages are Indo-European and high-resource with large Wikipedias). Importantly, they show that transfer effectiveness does not depend on the vocabulary overlap between the languages.   


%

In very recent work, concurrent to ours, \newcite{hu2020xtreme} introduce XTREME, a benchmark for evaluating multilingual encoders encompassing 9 tasks and 40 languages in total.\footnote{Note that individual tasks in XTREME do not cover all 40 languages, but rather significantly smaller language subsets.} Their primary focus is zero-shot transfer evaluation, while they also experiment with target-language fine-tuning on 1,000 instances for POS tagging and NER; this leads to substantial gains over zero-shot transfer. While \newcite{hu2020xtreme} focus on the evaluation aspects and protocols, in this work, we provide a more detailed analysis and understanding of the factors that hinder effective zero-shot transfer across diverse tasks. We also put more emphasis on few-shot learning scenarios, and approach it differently: we first fine-tune the MMTs on the (large) English task-specific training set and then fine-tune/adapt it further with a small number of target-language instances (e.g., even with as few as 10 instances).

\newcite{artetxe2019cross} and \newcite{wu2020emerging} have analyzed monolingual BERT models in different languages to explain transfer effectiveness of MMTs. Their main conclusion is that, similar as with static word embeddings, it is the topological similarities\footnote{\newcite{wu2020emerging} call it ``latent symmetries''. This is essentially the assumption of approximate (weak) isomorphism between monolingual (sub)spaces \cite{sogaard2018limitations}.} between the subspaces of individual languages captured by MMTs that enable effective cross-lingual transfer. For cross-lingual word embedding spaces, the assumption of approximate isomorphism does not hold for distant languages \cite{sogaard2018limitations,vulic2019we}, and in face of limited-size monolingual corpora \cite{vulic2020all}. In this work, we empirically demonstrate that the same is true for zero-shot transfer with MMTs: transfer performance substantially decreases as we extend our focus to more distant target languages with smaller pretraining corpora.   

\section{Zero-Shot Transfer: Analyses}
\label{s:zeroshot}
We first focus on Q1 and Q2 (see \S\ref{s:intro}): we conduct zero-shot language transfer experiments on five different tasks. We then analyze the drops with respect to linguistic (dis)similarities and sizes of pretraining corpora of target languages.


\subsection{Experimental Setup}\label{sec:exp}

\noindent \textbf{Tasks and Languages.} We experiment with -- \textbf{a)} lower-level structured prediction tasks: POS-tagging, dependency parsing, and NER and \textbf{b)} higher-level language understanding tasks: NLI and QA. The aim is to probe if the factors contributing to transfer performance differ between these two task groups. Across all tasks, we experiment with 21 languages in total. 

\vspace{1.4mm}

\noindent \textit{Dependency Parsing} (DEP). We use Universal Dependency treebanks~\citep[UD,][]{nivre_universal_2017} for English and following target languages (from 8 language families): Arabic (\textsc{ar}), Basque (\textsc{eu}), (Mandarin) Chinese (\textsc{zh}), Finnish (\textsc{fi}), Hebrew (\textsc{he}), Hindi (\textsc{hi}), Italian (\textsc{it}), Japanese (\textsc{ja}), Korean (\textsc{ko}), Russian (\textsc{ru}), Swedish (\textsc{sv}), and Turkish (\textsc{tr}).


\vspace{1.3mm}

\noindent \textit{Part-of-speech Tagging} (POS). Again, we use UD and obtain the Universal POS-tag (UPOS) annotations from the same treebanks as with DEP. 

\vspace{1.3mm}

\noindent \textit{Named Entity Recognition} (NER). We use the NER WikiANN dataset from \newcite{rahimi-etal-2019-massively}. We experiment with the same set of 12 target languages as in DEP and POS. 

\vspace{1.3mm}

\noindent \textit{Cross-lingual Natural Language Inference} (XNLI). We run our experiments on the XNLI corpus \cite{conneau2018xnli} created by crowd-translating the dev and test portions of the English Multi-NLI dataset \cite{N18-1101} into $14$ languages (French (\textsc{fr}), Spanish (\textsc{es}), German (\textsc{de}), Greek (\textsc{el}), Bulgarian (\textsc{bg}), Russian (\textsc{ru}), Turkish (\textsc{tr}), Arabic (\textsc{ar}), Vietnamese (\textsc{vi}), Thai (\textsc{th}), Chinese (\textsc{zh}), Hindi (\textsc{hi}), Swahili (\textsc{sw}), and Urdu (\textsc{ur})).  

\vspace{1.3mm}

\noindent \textit{Cross-lingual Question Answering} (XQuAD). We rely on the XQuAD dataset \cite{artetxe2019cross}, created by translating the 240 development paragraphs (from 48 documents) and their corresponding 1,190 question-answer pairs of SQuAD v1.1~\citep{rajpurkar2016} to $11$ languages (\textsc{es}, \textsc{de}, \textsc{el}, \textsc{ru}, \textsc{tr}, \textsc{ar}, \textsc{vi}, \textsc{th}, \textsc{zh}, and \textsc{hi}). Given a paragraph-question pair, the task is to identify the exact span in the paragraph, which contains the answer to the question. 
To enable the comparison between zero-shot and few-shot transfer (see \S\ref{sec:fewshot}), we reserve a portion of $10$ articles as the development set in our experiments and report the final performance on the remaining $38$ articles.

\vspace{1.6mm}
\noindent \textbf{Fine-tuning.} We adopt a standard fine-tuning-based approach for both mBERT and XLM-R in all tasks. Tokenization for both models is done in a standard fashion.\footnote{We tokenize the input for each of the two models with their accompanying pretrained fixed-vocabulary tokenizers: WordPiece tokenizer \cite{Wu:2016arxiv} with the vocabulary of 110K tokens for mBERT (we add special tokens \texttt{[CLS]} and \texttt{[SEP]}), and the SentencePiece BPE tokenizer \cite{sennrich2016neural} with the vocabulary of 250K tokens for XLM-R. Special tokens \texttt{<s>} and \texttt{</s>} are also added.} Particular classification architectures on top of the MMTs depend on the task: for DEP we use a biaffine dependency parser \cite{dozat_deep_2017}; for POS, we rely on a simple feed-forward token-level classifier; for NER, we feed MMT representations to a CRF-based classifier, similar to \citep{peters_semi-supervised_2017}. For XNLI, we add a simple softmax classifier taking the transformed representation of the sequence start token as input (\texttt{[CLS]} for mBERT, \texttt{<s>} for XLM-R); for XQuAD, we pool the transformed representations of all subword tokens as input to a span classification head -- a linear layer that computes the start and end of the answer span.

\vspace{1.6mm}
\noindent \textbf{Training and Evaluation Details.} 
We evaluate mBERT \textit{Base} \textit{cased}: $L=12$ transformer layers with the hidden state size of $H=768$, $A=12$ self-attention heads. In addition, we work with \textsc{XLM-R} \textit{Base}: $L=12$, $H=768$, $A=12$. 



For XNLI, we limit the input sequence length to $T = 128$ subword tokens and train in batches of $32$ instances. For XQuAD, we limit the input length of paragraphs to $T=384$ tokens and the length of questions to $Q=64$ tokens. We slide over paragraphs with a stride of $128$ and train in $12$-instance batches. For XNLI and XQuAD, we grid-search for the optimal learning rate $\lambda \in \{5 \cdot 10^{-5}, 3\cdot10^{-5}\}$, and number of training epochs $n \in \{2, 3\}$.
For DEP and POS, we fix the number of training epochs to $50$; for NER, we train for $10$ epochs. We train in batches of $128$ sentences, with maximal sequence length of $T = 512$ subword tokens. For all tasks we use Adam as the optimization algorithm \cite{kingma2015adam}. 

We report DEP performance in terms of Unlabeled Attachment Scores (UAS).\footnote{Using Labeled Attachment Score (LAS) would make the small differences in annotation schemes between languages a confounding factor for the analyses of transfer performance, impeding insights into relevant factors of the study: language proximity and size of target language corpora.} For POS, NER, and XNLI we report accuracy, and for XQuAD, we report the Exact Match (EM) score.




\subsection{Results and Preliminary Discussion}

\setlength{\tabcolsep}{1.9pt}
\begin{table*}[!t]\centering
\scriptsize
\begin{tabular}{lcaccccccccccccccccccccc}\toprule
& &\multirow{2}{*}{\textbf{\textsc{en}}} &\textbf{\textsc{zh}} &\textbf{\textsc{tr}} &\textbf{\textsc{ru}} &\textbf{\textsc{ar}} &\textbf{\textsc{hi}} &\textbf{\textsc{eu}} &\textbf{\textsc{fi}} &\textbf{\textsc{he}} &\textbf{\textsc{it}} &\textbf{\textsc{ja}} &\textbf{\textsc{ko}} &\textbf{\textsc{sv}} &\textbf{\textsc{vi}} &\textbf{\textsc{th}} &\textbf{\textsc{es}} &\textbf{\textsc{el}} &\textbf{\textsc{de}} &\textbf{\textsc{fr}} &\textbf{\textsc{bg}} &\textbf{\textsc{sw}}  & \textbf{\textsc{ur}} \\
 \textbf{Task} & \textbf{Model} & \multirow{-2}{*}{\textbf{\textsc{en}}}
 &\textbf{$\Delta$} &\textbf{$\Delta$} &\textbf{$\Delta$} &\textbf{$\Delta$} &\textbf{$\Delta$} &\textbf{$\Delta$} &\textbf{$\Delta$} &\textbf{$\Delta$} &\textbf{$\Delta$} &\textbf{$\Delta$} &\textbf{$\Delta$} &\textbf{$\Delta$} &\textbf{$\Delta$} &\textbf{$\Delta$} &\textbf{$\Delta$} &\textbf{$\Delta$} &\textbf{$\Delta$} &\textbf{$\Delta$} &\textbf{$\Delta$} &\textbf{$\Delta$} &\textbf{$\Delta$} \\\midrule
DEP& \textsc{B}  &92.3 &-40.9 &-41.2 &-23.5 &-47.9 &-49.6 &-42.0 &-26.7 &-29.7 &-10.6 &\textbf{-55.4} &-53.4 &-12.5 &- &- &- &- &- &- &- &- &- \\
POS&  \textsc{B} &95.5 &-33.6 &-26.6 &-9.5 &-32.8 &-33.9 &-28.3 &-14.6 &-21.4 &-6.0 &\textbf{-47.3} &-37.3 &-6.2 &- &- &- &- &- &- &- &- &- \\
NER & \textsc{B} &92.3 &-31.5 &-6.5 &-9.2 &-29.2 &-12.8 &-8.5 &-0.9 &-9.2 &-0.8 &\textbf{-51.1} &-12.9 &-1.9 &- &- &- &- &- &- &- &- &- \\ \midrule
\multirow{2}{*}{XNLI} &\textsc{B} &82.8 &-13.6 &-20.6 &-13.5 &-17.3 &-21.3 &- &- &- &- &- &- &- &-11.9 &-28.1 &-8.1 &-14.1 &-10.5 &-7.8 &-13.3 &\textbf{-33.0} &-23.4 \\
&\textsc{X} &84.3 &-11.0 &-11.3 &-9.0 &-13.0 &-14.2 &- &- &- &- &- &- &- &-9.7 &-12.3 &-5.8 &-8.9 &-7.8 &-6.1 &-6.6 &\textbf{-20.2} &-17.3 \\ \midrule
\multirow{2}{*}{XQuAD} &\textsc{B} &71.1 &-22.9 &-34.2 &-19.2 &-24.7 &-28.6 &- &- &- &- &- &- &- &-22.1 &\textbf{-43.2} &-16.6 &-28.2 &-14.8 &- &- &- &- \\
&\textsc{X} &72.5 &\textbf{-26.2} &-18.7 &-15.4 &-24.1 &-22.8 &- &- &- &- &- &- &- &-19.7 &-14.8 &-14.5 &-15.7 &-16.2 &- &- &- &- \\
\bottomrule
\end{tabular}
\caption{Zero-shot cross-lingual transfer performance on five tasks (DEP, POS, NER, XNLI, and XQuAD) with mBERT (B) and XLM-R (X). We show the monolingual \textsc{en} performance and report drops in performance relative to \textsc{en} for all target languages. Numbers in bold indicate the largest zero-shot performance drops for each task.}\label{tab:zero}
\end{table*}

A summary of the zero-shot cross-lingual transfer results per target language is provided in Table~\ref{tab:zero}. 
Unsurprisingly, we observe substantial drops in performance for all tasks and all target languages compared to the reference \textsc{en} performance. However, the extent of decrease varies greatly across languages. For instance, NER transfer with mBERT for \textsc{it} drops a mere 0.8\%, whereas it is 51\% on \textsc{ja} NER. On XNLI, transferring with XLM-R yields a moderate decrease of 6.1\% for \textsc{fr}, but a much larger drop of 20\% for \textsc{sw}.       
%
%
At first glance, it would appear -- as suggested in prior work -- that the drops in transfer performance primarily depend on the language (dis)similarity, and that they are much more pronounced for languages which are more distant from \textsc{en} (e.g., \textsc{ja}, \textsc{zh}, \textsc{ar}, \textsc{th}, \textsc{sw}). 

While we do not observe a notable exception to this pattern on the three lower-level tasks, language proximity alone does not explain many results obtained on XNLI and XQuAD. For instance, on XNLI (for both mBERT and XLM-R), the \textsc{ru} scores are comparable to those on \textsc{zh}, while they are lower for \textsc{hi} and \textsc{ur}: this is despite the fact that as Indo-European languages \textsc{ru}, \textsc{hi}, and \textsc{ur} are linguistically closer to \textsc{en} than \textsc{zh}. Similarly, we observe comparable scores on XQuAD for \textsc{th}, \textsc{ru}, and \textsc{es}. Therefore, in what follows we seek a more informed explanation of the obtained zero-shot transfer results.     

\subsection{Analysis}
For each task, we now analyze the correlations between transfer performance in the task and \textbf{a)} various measures of linguistic proximity (i.e., similarity) between languages and \textbf{b)} the size of MMT pretraining corpora of each target language. 
\vspace{1.6mm}

\noindent \textbf{Language Vectors and Corpora Sizes.} In order to estimate linguistic similarity, we rely on language vectors from \textsc{lang2vec}, which encode various linguistic features from the URIEL database \cite{littell-etal-2017-uriel}. We consider several \textsc{lang2vec}-based language vectors as follows: \texttt{syntax} (\textsc{SYN}) vectors encode syntactic properties, e.g., whether a subject appears before or after a verb; \texttt{phonology} (\textsc{PHON}) vectors encode phonological properties of a language, e.g. the consonant-vowel ratio; \texttt{inventory} (\textsc{INV}) vectors encode presence or absence of natural classes of sounds, e.g., voiced uvulars; FAM vectors denote memberships in \texttt{language families}, e.g., Indo-Germanic; and GEO vectors express orthodromic distances for languages w.r.t. a fixed number of points on the Earth’s surface. 
%
Language similarity is then computed as the cosine similarity between the languages' corresponding \textsc{lang2vec} vectors. Each aspect listed above (e.g., SYN, GEO, FAM) yields one scalar feature for our analysis.   

We also include another feature: the z-normalized size of the
target language corpus used in MMT model pretraining (\textsc{SIZE}).\footnote{For XLM-R, we take the reported sizes of language-specific portions of CommonCrawl-100 from \newcite{Conneau:2020acl}; for mBERT, we take the sizes of language-specific Wikipedias.\ We take the sizes of Wikipedia snapshots from October 21, 2018. While these are not the exact snapshots on which mBERT was pretrained, the size ratios between languages should be roughly the same.} 


\setlength{\tabcolsep}{7pt}
\begin{table*}[t!]
\centering
\small
\begin{tabularx}{\linewidth}{l c cc cc cc cc cc cc} 
\toprule
 &  & \multicolumn{2}{c}{\textsc{SYN}} & \multicolumn{2}{c}{\textsc{PHON}} & \multicolumn{2}{c}{\textsc{INV}} & \multicolumn{2}{c}{\textsc{FAM}} & \multicolumn{2}{c}{\textsc{GEO}} & \multicolumn{2}{c}{\textsc{SIZE}}  \\ \cmidrule(lr){3-4}\cmidrule(lr){5-6}\cmidrule(lr){7-8}\cmidrule(lr){9-10}\cmidrule(lr){11-12}\cmidrule(lr){13-14}
\textbf{Task} & \textbf{Model} & P & S & P & S & P & S & P & S & P & S & P & S \\ \midrule
DEP & mBERT & \textbf{0.93} & \textbf{0.92} & 0.80 & 0.81 & 0.51 & 0.02 & 0.73 & 0.63 & 0.70 & 0.71 & 0.75 & 0.55 \\ 
POS & mBERT & \textbf{0.89} & \textbf{0.87} & 0.84 & 0.84 & 0.43 & -0.04 & 0.66 & 0.63 & 0.78 & 0.81 & 0.65 & 0.50 \\ 
NER & mBERT & 0.56 & 0.62 & \textbf{0.78} & \textbf{0.86} & 0.23 & 0.02 & 0.39 & 0.49 & \textbf{0.75} & \textbf{0.88} & 0.27 & 0.23 \\ \midrule
\multirow{2}{*}{XNLI} & XLM-R & \textbf{0.88} & \textbf{0.90} & 0.29 & 0.27 & 0.31 & -0.11 & 0.63 & 0.54 & 0.54 & 0.74 & 0.70 & 0.76 \\
& mBERT & \textbf{0.87} & 0.86 & 0.21 & 0.08 & 0.29 & 0.04 & 0.61  & 0.47 & 0.55 & 0.67 & 0.77 & \textbf{0.91} \\ \midrule
\multirow{2}{*}{XQuAD} & XLM-R & 0.69 & 0.53 & \textbf{0.85} & \textbf{0.81} & 0.62 & -0.01 & \textbf{0.81} & 0.54 & 0.43 & 0.50 & \textbf{0.81} & 0.55 \\
& mBERT & 0.84 & 0.89 & 0.56 & 0.48 & 0.55 & 0.22 & 0.79 & 0.64 & 0.51 & 0.55  & \textbf{0.89} & \textbf{0.96} \\
\bottomrule
\end{tabularx}
\caption{Correlations between zero-shot transfer performance with mBERT and XLM-R for different downstream tasks, across a set of target languages, with linguistic proximity scores (features \textsc{SYN}, \textsc{PHON}, \textsc{INV}, \textsc{FAM} and \textsc{GEO}) and the pretraining size of the target language corpora (feature \textsc{SIZE}). The results are reported in terms of Pearson (P) and Spearman (S) correlation coefficients. Highest correlations for each task-model pair are in bold.
}
\vspace{-1.5mm}
\label{tab:corr}
\end{table*}

\vspace{1.6mm}
\noindent \textbf{Correlation Analysis.} 
We first correlate individual features with the zero-shot transfer scores for each task and show the results in Table~\ref{tab:corr}.
Quite intuitively, the zero-shot scores for low-level syntactic tasks -- POS and DEP -- best correlate with syntactic similarity (\textsc{SYN}). \textsc{SYN} similarity also correlates quite highly with transfer results for higher-level tasks, except for XLM-R on XQuAD. Phonological similarity (\textsc{PHON}) correlates best with the transfer results of mBERT on NER and XLM-R on XQuAD. 
Interestingly, for both high-level tasks and both models, we observe very high correlations between transfer performance and the size of pretraining corpora of the target language (\textsc{SIZE}). On the other hand, \textsc{SIZE} shows substantially lower correlations with transfer performance across lower-level tasks (DEP, POS, NER). We believe that this is because language understanding tasks such as NLI and QA require rich representations of the semantic phenomena of a language, whereas low-level tasks require simpler structural knowledge of the language -- to acquire the former with distributional models, one simply needs much more text than to acquire the latter.







\vspace{1.6mm}
\noindent \textbf{Meta-Regression.} 
 We observe high correlations between the transfer scores and several individual features (e.g., \textsc{SYN}, \textsc{PHON} and \textsc{SIZE}). Therefore, we further test if even higher correlations can be achieved by a (linear) combination of the individual features. For each task, we fit a linear regression using target language performances on the task as labels. For each task, we have between 11 and 14 target languages (i.e., instances for fitting the linear regression); we thus evaluate the regressor's performance, i.e., a Pearson correlation score between the learned linear combination of features and transfer performance, via leave-one-out cross-validation (LOOCV). In order to allow for only a subset of most useful features to be selected, we perform greedy forward feature selection: we start from an empty feature set and in each iteration add the feature that boosts LOOCV performance the most; we stop when none of the remaining features further improve the Pearson correlation.   
%

%
%
\setlength{\tabcolsep}{2.2pt}
\begin{table}[t!]
\centering
\small
\begin{tabularx}{\linewidth}{l c l ccc}
\toprule
\textbf{Task} &\textbf{Model} & \textbf{Selected features} & P & S & MAE \\ \midrule
POS & B & \textsc{SYN}\,(.99) & 0.94 & 0.90 & 4.60 \\
DEP & B & \textsc{SYN}(.99) & 0.93 & 0.92 & 5.77 \\
NER & B & \textsc{PHON}(.99) & 0.69 & 0.82 & 9.45 \\ \midrule 
\multirow{2}{*}{\textsc{XNLI}} & X & \textsc{SYN}\,(.51);\,\textsc{SIZE}\, (.49) & 0.84 & 0.85 & 2.01\\
& \multirow{2}{*}{\textsc{B}} & \textsc{SYN}\,(.35);\,\textsc{SIZE}\,(.34),  & \multirow{2}{*}{0.89} & \multirow{2}{*}{0.90} & \multirow{2}{*}{2.78} \\
& &  \textsc{FAM}\,(.31) & & & \\
\midrule
\multirow{2}{*}{XQuAD} & X & \textsc{PHON}\,(.99) 
& 0.95 & 0.83 & 2.89 \\
& B & \textsc{SIZE}\,(.99) & 0.89 & 0.93 & 4.76 \\
\bottomrule
\end{tabularx}
\caption{Results of the (linear) meta-regressor: predicting zero-shot transfer performance with mBERT\,(B) and XLM-R\,(X), for each of our five tasks, from the set of features indicating language proximity to the source language (\textsc{en}) and the size of the target language corpora in pretraining MMT pretraining. We list only the features with assigned weights $\geq 0.01$. P=Pearson; S=Spearman; MAE=Mean Average Error.}\label{tab:meta}
\end{table}

The ``meta-regression'' results are summarized in Table~\ref{tab:meta}. For each task-model pair, we list the features selected with the greedy feature selection and show (normalized) weights assigned to each feature.
%
Except for NER, a linear combination of features yields higher correlations with the zero-shot transfer results than any of the individual features. These results empirically confirm our intuitions from the previous section that (structural) linguistic proximity (\textsc{SYN}) explains zero-short transfer performance for the low-level structural tasks (DEP and POS), but that it cannot fully explain performance in the two language understanding tasks. For XNLI, the transfer results are best explained with the combination of structural language proximity (\textsc{SYN}) and the size of the target-language pretraining corpora (SIZE). For XQuAD with mBERT, \textsc{SIZE} alone best explains zero-short transfer scores. We note that the features are also mutually very correlated (e.g., languages closer to \textsc{en} tend to have larger corpora): if the regressor selects only one feature, this does not suggest that other features do not correlate with transfer results (see Table~\ref{tab:corr}).


The coefficients in Table~\ref{tab:meta} again indicate the importance of SIZE for the language understanding tasks and highlight our core finding: pretraining corpora sizes are stronger features for predicting zero-shot performance in higher-level tasks, whereas the results in lower-level tasks are more affected by typological language proximity. 

\section{From Zero to Hero: Few-Shot}
\label{sec:fewshot}
Motivated by the low zero-shot results across many tasks and languages in \S\ref{s:zeroshot}, we now investigate Q4 from \S\ref{s:intro}, aiming to mitigate the transfer gap by relying on inexpensive few-shot transfer settings with a small number of target--language examples.

\vspace{1.6mm}
\noindent \textbf{Experimental Setup.} We rely on the same models, tasks, and evaluation protocols as described in \S\ref{sec:exp}. However, instead of fine-tuning the model on task-specific data in \textsc{en} only, we continue the fine-tuning process by feeding $k$ additional training examples randomly chosen from reserved target language data portions, disjoint with the test sets.\footnote{Note that for XQuAD, we performed the split on the article level to avoid topical overlap. Consequently, $k$ there refers to the number of articles.} For our lower-level tasks, we compare three sampling methods: (i) random sampling (\textsc{rand}) of $k$ target language sentences, (ii) selection of the $k$ shortest (\textsc{shortest}) and (iii) the $k$ longest (\textsc{longest}) sentences. 
In all three cases, we only choose between sentences with $\geq 3$ and $\leq 50$ tokens. 
For XNLI and XQuAD, we run the experiments five times and report the average score.

\subsection{Results and Discussion}
\label{ss:rdfew}

\setlength{\tabcolsep}{4.5pt}
\begin{table*}[!t]\centering
\small
\begin{tabularx}{\linewidth}{lllarrrrrrrrrrr}\toprule
\rowcolor{white}
& & & $k$ &\multicolumn{2}{c}{$k=10$} &\multicolumn{2}{c}{$k=50$} &\multicolumn{2}{c}{$k=100$} &\multicolumn{2}{c}{$k=500$} &\multicolumn{2}{c}{$k=1000$} \\ 

\cmidrule(lr){5-6}
\cmidrule(lr){7-8}
\cmidrule(lr){9-10}
\cmidrule(lr){11-12}
\cmidrule(lr){13-14}
 Task &Model &Sampling & \cellcolor{white} $k=0$ &score &$\Delta$ &score &$\Delta$ &score &$\Delta$ &score &$\Delta$ &score &$\Delta$ \\\midrule
\multirow{3}{*}{DEP} &\multirow{3}{*}{\textsc{mBERT}} &Random &59.00 &69.32 &10.32 &75.20 &16.20 &77.32 &18.32 &82.26 &23.26 &84.32 &25.32 \\
& &Shortest &59.00 &59.34 &0.34 &61.12 &2.12 &63.25 &4.25 &72.94 &13.94 &76.82 &17.82 \\
& &Longest &59.00 &73.45 &14.45 &78.45 &19.45 &80.04 &21.04 &84.16 &25.16 &85.73 &26.73 \\
\cmidrule{3-14}
\multirow{3}{*}{POS} &\multirow{3}{*}{\textsc{mBERT}} &Random &72.65 &81.69 &9.04 &86.45 &13.80 &88.10 &15.45 &91.75 &19.10 &93.07 &20.42 \\
& &Shortest &72.65 &73.51 &0.86 &77.81 &5.16 &81.90 &9.25 &87.91 &15.26 &90.36 &17.71 \\
& &Longest &72.65 &86.07 &13.42 &89.63 &16.98 &90.94 &18.29 &93.42 &20.77 &94.21 &21.56 \\
\cmidrule{3-14}
\multirow{3}{*}{NER} &\multirow{3}{*}{\textsc{mBERT}} &Random &78.33 &86.13 &7.80 &89.11 &10.78 &90.04 &11.71 &91.48 &13.15 &92.44 &14.11 \\
& &Shortest &78.33 &76.04 &-2.29 &75.49 &-2.84 &76.97 &-1.36 &76.93 &-1.40 &80.19 &1.86 \\
& &Longest &78.33 &83.30 &4.97 &84.33 &6.00 &84.98 &6.65 &87.49 &9.16 &88.88 &10.55 \\
\midrule
\multirow{2}{*}{XNLI} &\textsc{mBERT} &Random &65.92 &65.89 &-0.03 &65.08 &-0.84 &64.92 &-1.00 &67.41 &1.49 &68.16 &2.24 \\
&\textsc{XLM-R} &Random &73.32 &73.73 &0.41 &73.76 &0.45 &75.03 &1.71 &75.34 &2.02 &75.84 &2.52 \\
\cmidrule(lr){3-14}
\rowcolor{white}
& & & & \multicolumn{2}{c}{$k=2$} & \multicolumn{2}{c}{$k=4$} &\multicolumn{2}{c}{$k=6$} &\multicolumn{2}{c}{$k=8$} & \multicolumn{2}{c}{$k=10$} \\
\cmidrule(lr){5-6}
\cmidrule(lr){7-8}
\cmidrule(lr){9-10}
\cmidrule(lr){11-12}
\cmidrule(lr){13-14}
\multirow{2}{*}{XQUAD} &\textsc{mBERT} &Random &45.62 &48.12 &2.50 &48.66 &3.04 &49.34 &3.72 &49.91 &4.29 &50.19 &4.57 \\
&\textsc{XLM-R} &Random &53.68 &53.73 &0.05 &53.84 &0.17 &54.76 &1.08 &55.56 &1.88 &55.78 &2.10 \\
\bottomrule
\end{tabularx}
\caption{Results of the few-shot experiments with varying numbers of target-language examples $k$. For each $k$, we report performance averaged across languages and the difference ($\Delta$) with respect to the zero-shot setting.}\label{tab:res_few_shot}
\end{table*}

\begin{figure*}[ht!]
    \centering
    \includegraphics[width=0.85\textwidth]{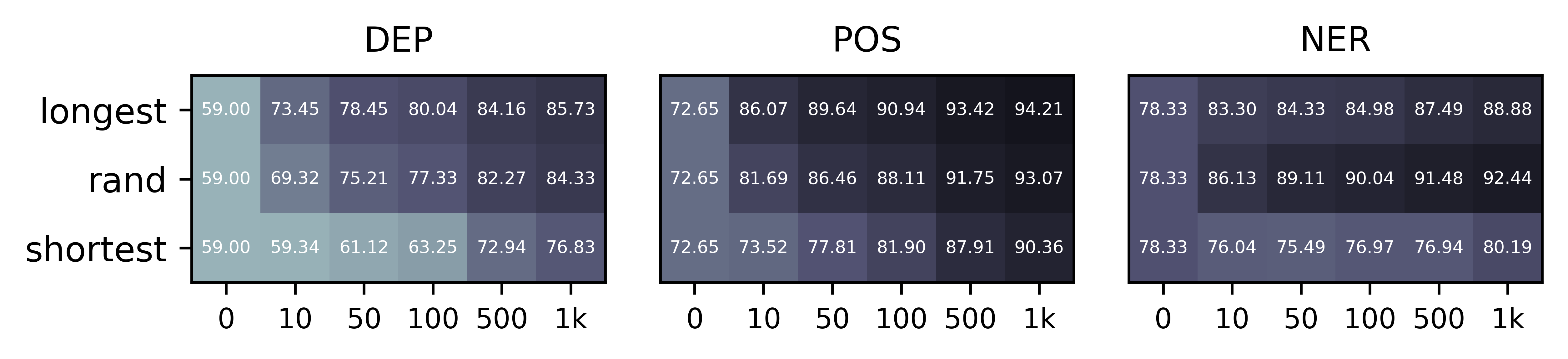}
    \vspace{-2mm}
    \caption{Heatmaps showing trends in performance improvement with few-shot data augmentation; the X-axis indicates the number of target-language instances $k$, whereas the Y-axis indicates the choice of sampling method.}
    \label{fig:heatmap}
\end{figure*}{}

The results on each task, conditioned on the number of examples $k$ and averaged across all target languages, are presented in Table~\ref{tab:res_few_shot}. We note substantial improvements in few-shot learning setups for all tasks. However, the results also reveal notable differences between different types of tasks. For higher-level language understanding tasks the improvements are less pronounced; the maximum gains after seeing $k=1,000$ target-language instances and $10$ articles, respectively, are between $2.1$ and $4.57$ points. On the other hand, the gains for the lower-level tasks are massive: between $14.11$ and $26$ percentage points, choosing the best sampling strategy for each task. Moreover, the gains in all lower-level tasks are substantial even when we add only 10 annotated sentences in the target language (on average, up to 14.45 percentage points on DEP and 13.42 points on POS). What is more, additional experiments (omitted for brevity) show substantial gains for DEP and POS even when we add fewer than 5 annotated sentences.

A comparison of different sampling strategies for the lower-level tasks is also shown in Figure~\ref{fig:heatmap}. 
For DEP and POS, the pattern is very clear and very expected -- adding longer sentences results in better scores. For NER, however, \textsc{rand} appears to perform best, with a larger gap between \textsc{rand} and \textsc{smallest}. We hypothesize that this is due to very long sentences being relatively sparse with named entities, resulting in our model seeing a lot of negative examples; shorter sentences are also less helpful than for DEP and POS because they consist of (confirmed by inspection) a single named entity mention, without non-NE tokens.

\begin{figure*}[ht!]
    \begin{subfigure}{0.49\textwidth}
        \includegraphics[width=\textwidth]{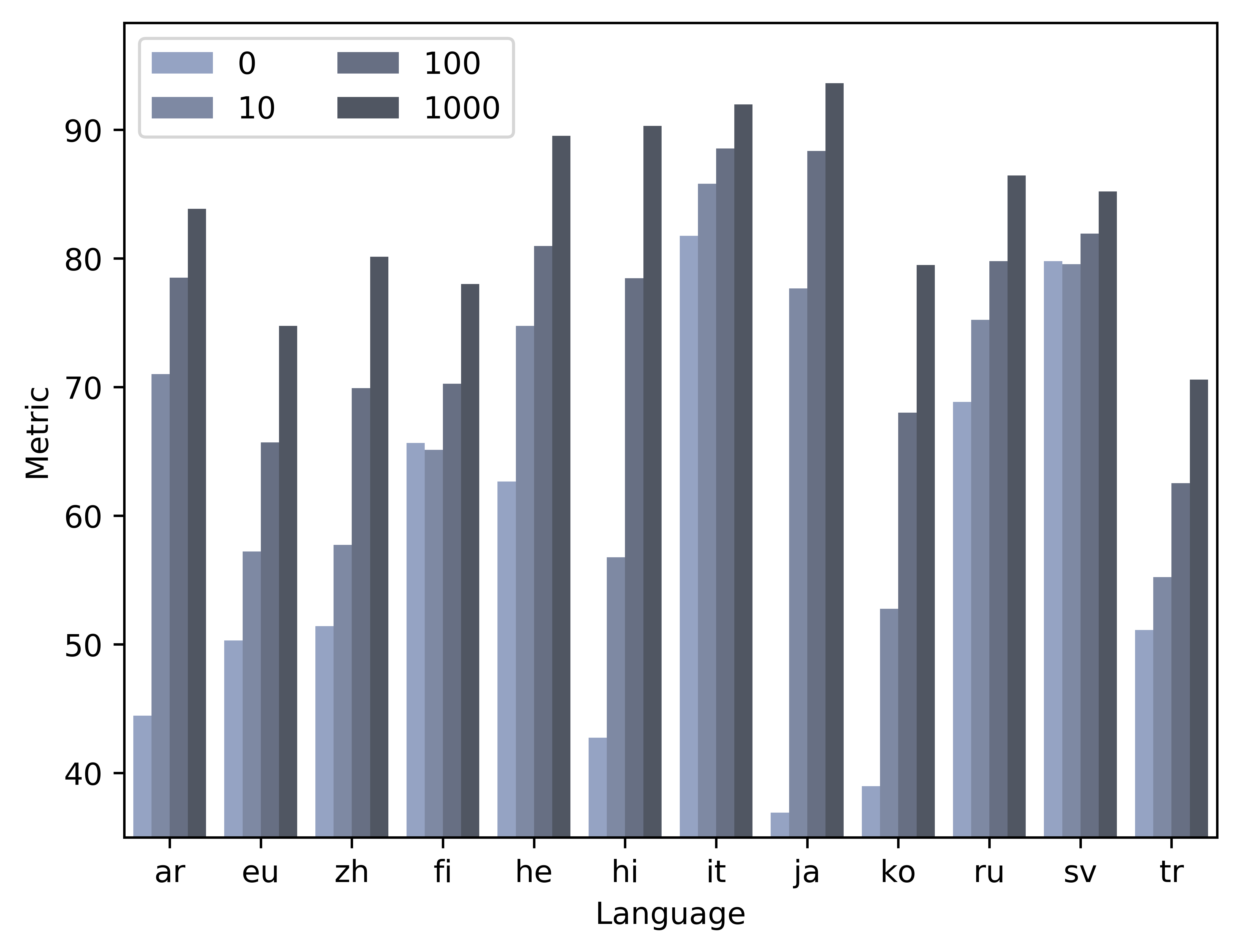}
        \caption{DEP}
    \end{subfigure}
    \begin{subfigure}{0.49\textwidth}
        \includegraphics[width=\textwidth]{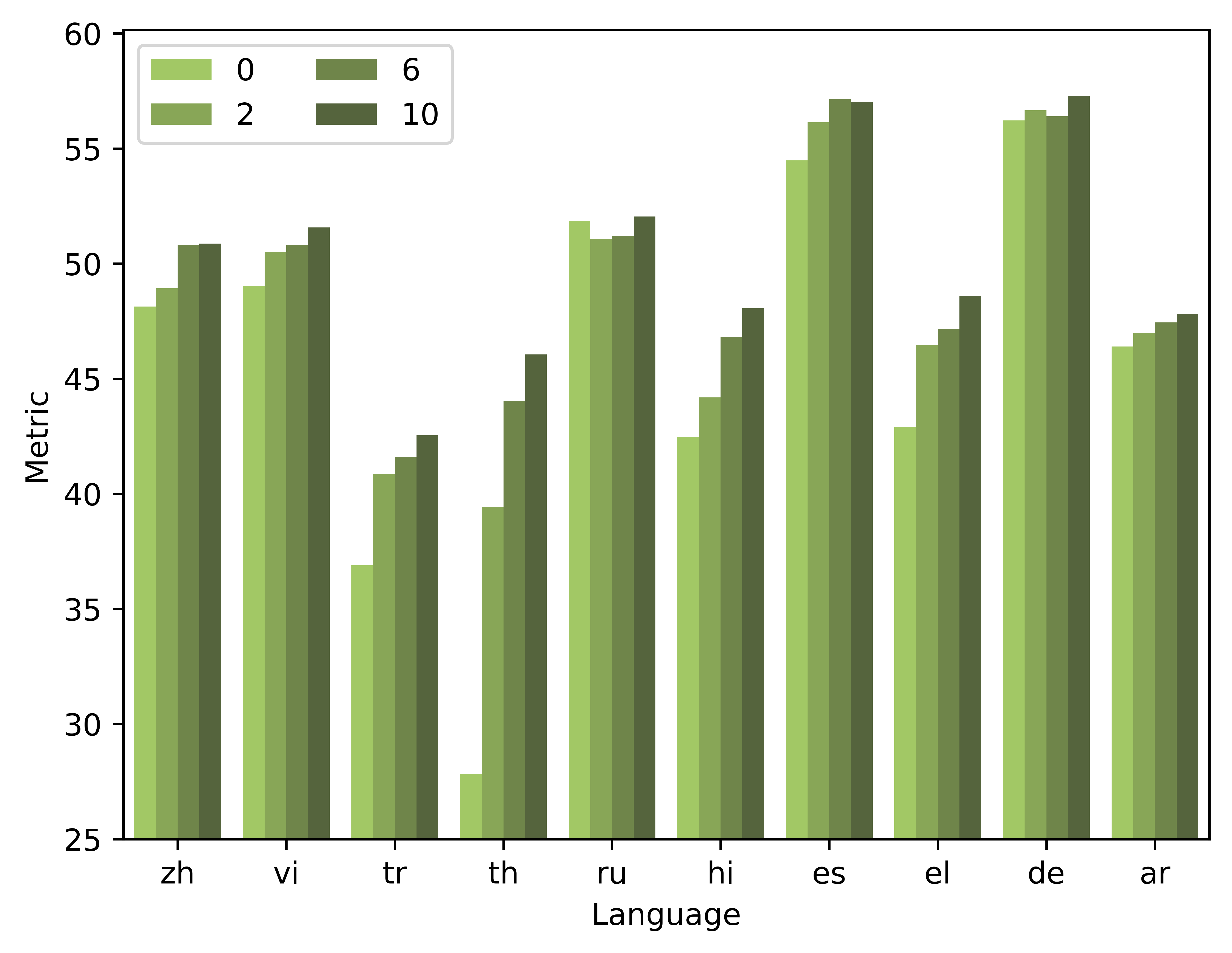}
        \caption{XQuAD}
    \end{subfigure}
    \caption{Few-shot results for each language with varying $k$ for a) DEP, reported in terms of UAS score, and b) XQuAD accuracy. For DEP, $k$ corresponds to the number of sampled sentences, and for XQuAD, to the number of sampled articles. For the other tasks, we refer the reader to Appendix~\ref{sec:app_gains}.}
    \label{fig:bar_all}
\end{figure*}

Figure \ref{fig:bar_all} illustrates few-shot performance for individual languages on one lower-level (DEP) and one higher-level task (XQuAD), for different values of $k$.\footnote{We provide the analogous figures for the remaining three tasks in the Appendix.} 
%
%
Across languages, we see a clear trend -- more distant target languages benefit much more from the few-shot data. Observe, for instance, \textsc{sv} (DEP, a) or \textsc{de} (XQuAD, b). Both are closely related to \textsc{en}, both generally have high scores in the zero-shot transfer, and both benefit only marginally from few-shot data points. We suspect that for such closely related languages, with enough MMT pretraining data, the model is able to extract missing knowledge from a few language-specific data points; its priors for languages closer to \textsc{en} are already quite sensible and \textit{a priori} have a smaller room for improvement. 
In stark contrast, \textsc{ko} (DEP, a) and \textsc{th} (XQuAD, b), both exhibit fairly poor zero-shot performance and understandably so, given their linguistic distance to \textsc{en}. 
Given in-language data, however, both see rapid leaps in performance, displaying a gain of almost 40\% UAS on DEP, and almost 5\% on XQuAD. In a sense, this can be seen as the models' ability to rapidly learn to utilize the multilingual representation space to adapt its knowledge of the downstream task to the target language. 

Other interesting patterns emerge. Particularly interesting are DEP results for \textsc{ja} and \textsc{ar}, where we observe massive UAS improvements with only 10 annotated sentences. For XQuAD, we observe a substantial improvement from only $2$ in-language documents for \textsc{th}. 
In sum, we see the largest gains from few-shot transfer exactly on languages where the zero-shot transfer setup yields the worst performance: languages distant from \textsc{en} and represented with a small corpus in MMT pretraining.   

{}


\subsection{Further Discussion}
As the results in \S\ref{ss:rdfew} prove, moving to the few-shot transfer setting can substantially improve performance and reduce the gaps observed with zero-shot transfer, especially for lower-resourced languages. While additional fine-tuning of MMT on the small number of target-language instances is computationally cheap, the potential bottleneck lies in possibly expensive data annotation. This is, especially for minor languages, potentially a major issue and deserves further analysis. What are the annotation costs, and at which conversion rate do they translate into performance points? 
Here, we provide some ballpark estimates based on annotation costs reported by other researchers.

\vspace{1.3mm}

\noindent \textbf{Natural Language Inference.} \newcite{marelli-etal-2014-sick} reportedly paid \$$2,030$ for $200$k judgements, which would amount to \$$0.01015$ per NLI instance and, in turn, to \$$10.15$ for $1,000$ annotations. In our few-shot experiments this would yield an average improvement of $2.24$ and $2.54$ accuracy points for mBERT and XLM-R, respectively.

\vspace{1.3mm}

\noindent \textbf{Question Answering. }
\citet{rajpurkar2016} report a payment cost of \$$9$ per hour and a time effort of $4$ minutes per paragraph. With an average of $5$ paragraphs per article, our few-shot scenario ($10$ articles) roughly requires $50$ paragraphs-level annotations, i.e., $200$ minutes of annotation effort and would in total cost around \$$30$ (for respective performance improvements of 4.5 and 2.1 points for mBERT and XLM-R). 


On the one hand, compared to language understanding tasks, our lower-level (DEP, POS) tasks are presumably more expensive to annotate, as they require some linguistic knowledge and annotation training. On the other hand, as shown in our few-shot experiments, we typically need much fewer annotated instances (i.e., we observe high gains with already $10$ target language sentences) for substantial gains in these tasks.   

\vspace{1.3mm}

\noindent \textbf{Dependency Parsing.} \citet{tratz_dependency_2019} provide an overview of crowd-sourcing annotations for dependency parsing; they report obtaining
a fully correct dependency tree from at least one annotator for 72\% of sentences.
At the reported cost of \$0.28 per sentence this amounts to spending \$280 for annotating $1,000$ sentences. Somewhat shockingly, annotating $10$ sentences with dependency trees -- which for some languages like \textsc{ar} and \textsc{ja} corresponds to performance gains of 30-40 UAS points (see Figure \ref{fig:bar_all}) -- amounts to spending merely \$3-5.    

\vspace{1.3mm}

\noindent \textbf{Part-of-Speech Tagging.} \citet{hovy_experiments_2014} measure agreement of crowdsourced {POS} annotations with expert annotations; they crowdsource annotations for 1,000 tweets, at a cost of \$0.05 for every 10 tokens. With a total of $14,619$ tokens in the corpus, this amounts to approximately \$73 for $1,000$ tweets, which is $\geq 1,000$ sentences.\footnote{Note, however, that lower-level tasks do come with an additional risk of poorer quality annotation, due to crowdsourced annotators not being experts. \newcite{Garrette:2013naacl} report that even for truly low-resource languages (e.g., Kinyarwanda, Malagasy), it is possible to obtain $\approx$ 100 POS-annotated sentences.} Based on Table~\ref{tab:res_few_shot}, 2 hours of POS annotation work translates to gains of up to 18 points on average over zero-shot transfer methods. 


\vspace{1.3mm}

\noindent \textbf{Named Entity Recognition.} \citet{ide_crowdsourcing_2017} provide estimates for crowdsourcing annotation for {named entity recognition}; they pay \$0.06 per sentence, resulting in \$60 cost for $1,000$ annotated sentences. At a median pay of \$11.37/hr, this amounts to around 190 sentences annotated in an hour. In other words, in less than 3 hours, we can collect more than 500 annotated examples. According to Table~\ref{tab:res_few_shot}, this can result in gains of 10-14 points on average, and even more for some languages (e.g., 27 points for \textsc{ar}).

A provocative high-level question that calls for further discussion in future work can be framed as: are GPU hours effectively more costly\footnote{financially, but also ecologically~\citep{strubell2019energy}.} than data annotations are in the long run? While MMTs are extremely useful as general-purpose models of language, their potential for some (target) languages can be quickly unlocked by pairing them with a small number of annotated target-language examples. Effectively, this suggests leveraging the best of both worlds, i.e., coupling knowledge encoded in large MMTs with a small annotation effort.

\section{Conclusion}
Research on zero-shot language transfer in NLP is motivated by inherent data scarcity: the fact that most languages have no annotated data for most NLP tasks. Massively multilingual transformers (MMTs) have recently been praised for their zero-shot transfer capabilities that mitigate the data scarcity issue. In this work we have demonstrated that, similar to earlier language transfer paradigms, MMTs perform poorly in zero-shot transfer to distant target languages, and for languages with smaller monolingual data for pretraining. We have presented a detailed empirical analysis of factors affecting transfer performance of MMTs across diverse tasks and languages. Our results have revealed that structural language similarity determines the transfer success for lower-level tasks like POS-tagging and parsing; on the other hand, the pretraining corpora size of the target language is crucial for explaining transfer results for higher-level language understanding tasks. Finally, we have shown that the MMT potential on distant and lower-resource target languages can be quickly unlocked if they are offered a handful of annotated target-language instances. This finding provides strong evidence towards intensifying future research efforts focused on the more effective few-shot learning setups.

\section*{Acknowledgements}
This work is supported by the Eliteprogramm of the Baden-W\"{u}rttemberg Stiftung (AGREE Grant). The work of Ivan Vuli\'{c} is supported by the ERC Consolidator Grant LEXICAL (no 648909).
%
%
%



\bibliography{references}
\bibliographystyle{acl_natbib}

\clearpage
\appendix

\onecolumn
\section{Full data}
\label{sec:appendix}

\begin{table*}[ht!]

\resizebox{\textwidth}{!}{%
\begin{tabular}{lrrrrrrrrrrrrr}
\toprule
POS & ar        & eu        & zh        & fi        & he        & hi        & it        & ja        & ko        & ru        & sv        & tr        \\
\midrule
0     & 62.76 & 67.25 & 61.92 & 80.93 & 74.11 & 61.62 & 89.58 & 48.27 & 58.27 & 86.00 & 89.32 & 68.89 \\
10    & 83.58 & 76.18 & 74.23 & 82.07 & 81.06 & 79.20 & 90.48 & 81.67 & 73.54 & 85.60 & 88.41 & 71.09 \\
50    & 90.12 & 82.30 & 83.98 & 82.79 & 89.80 & 84.14 & 93.65 & 89.28 & 75.02 & 88.54 & 91.18 & 77.92 \\
100   & 91.50 & 83.83 & 85.61 & 85.00 & 89.90 & 86.53 & 93.87 & 90.97 & 81.29 & 89.83 & 92.13 & 79.61 \\
500   & 94.66 & 87.72 & 89.78 & 88.83 & 94.83 & 90.34 & 95.75 & 93.76 & 87.34 & 93.75 & 93.55 & 87.56 \\
1000  & 95.35 & 88.65 & 91.40 & 90.79 & 95.85 & 93.39 & 96.30 & 94.28 & 90.29 & 94.31 & 94.86 & 89.00\\
\bottomrule
\end{tabular}%
}

\resizebox{\textwidth}{!}{%
\begin{tabular}{lrrrrrrrrrrrrr}
\toprule
NER & ar        & eu        & zh     & fi        & he        & hi        & it        & ja        & ko        & ru        & sv        & tr        \\
\midrule
0     & 63.16 & 83.81 & 60.87 & 91.44 & 83.16 & 79.56 & 91.51 & 41.21 & 79.39 & 83.08 & 90.43 & 85.86 \\
10    & 82.70 & 93.44 & 82.17 & 92.33 & 82.55 & 79.56 & 92.46 & 76.28 & 81.36 & 85.29 & 93.88 & 91.73 \\
50    & 86.94 & 94.37 & 86.83 & 93.12 & 86.99 & 85.57 & 92.83 & 78.53 & 86.27 & 89.52 & 96.02 & 92.06 \\
100   & 87.66 & 95.12 & 87.71 & 93.61 & 87.67 & 86.58 & 93.26 & 80.90 & 89.21 & 90.32 & 96.02 & 92.65 \\
500   & 90.46 & 95.36 & 90.44 & 94.55 & 91.03 & 88.17 & 94.03 & 81.50 & 91.44 & 91.17 & 97.09 & 94.40 \\
1000  & 90.46 & 95.95 & 91.27 & 94.31 & 91.52 & 90.44 & 94.96 & 85.63 & 92.17 & 92.84 & 97.33 & 94.39\\
\bottomrule
\end{tabular}%
}

\resizebox{\textwidth}{!}{%
\begin{tabular}{lrrrrrrrrrrrrr}
\toprule
DEP & ar        & eu        & zh         & fi        & he        & hi        & it        & ja        & ko        & ru        & sv        & tr        \\
\midrule
0     & 44.46 & 50.31 & 51.41 & 65.66 & 62.65 & 42.75 & 81.76 & 36.93 & 38.98 & 68.85 & 79.79 & 51.11 \\
10    & 71.00 & 57.23 & 57.73 & 65.13 & 74.75 & 56.76 & 85.80 & 77.67 & 52.76 & 75.23 & 79.55 & 55.22 \\
50    & 75.84 & 63.99 & 66.73 & 69.26 & 79.45 & 72.84 & 88.10 & 85.75 & 63.76 & 77.95 & 81.89 & 59.73 \\
100   & 78.50 & 65.70 & 69.91 & 70.25 & 80.98 & 78.47 & 88.54 & 88.35 & 68.00 & 79.78 & 81.93 & 62.54 \\
500   & 82.73 & 72.37 & 77.19 & 75.04 & 87.99 & 85.99 & 90.63 & 91.90 & 76.55 & 84.59 & 84.51 & 67.71 \\
1000  & 83.85 & 74.75 & 80.12 & 78.00 & 89.54 & 90.31 & 91.97 & 93.63 & 79.49 & 86.45 & 85.21 & 70.58\\
\bottomrule
\end{tabular}%
}

\resizebox{\textwidth}{!}{%
\begin{tabular}{lrrrrrrrrrrrrrrr}
\toprule
XNLI & fr    & es    & el    & bg    & ru    & tr    & ar    & vi    & th    & zh    & hi    & sw    & ur    & de    \\
\midrule
0     & 75.05 & 74.71 & 68.68 & 69.50 & 69.34 & 62.18 & 65.53 & 70.88 & 54.69 & 69.26 & 61.50 & 49.84 & 59.38 & 72.34 \\
10    & 75.09 & 73.62 & 67.04 & 69.35 & 69.80 & 61.86 & 65.56 & 69.26 & 55.30 & 70.89 & 61.92 & 51.79 & 59.28 & 71.63 \\
50    & 74.60 & 73.91 & 66.44 & 68.37 & 69.05 & 60.99 & 64.63 & 70.29 & 51.17 & 71.32 & 60.08 & 49.95 & 58.83 & 71.43 \\
100   & 73.85 & 73.50 & 65.67 & 68.47 & 70.24 & 60.13 & 64.93 & 69.59 & 51.68 & 71.46 & 60.01 & 48.96 & 58.78 & 71.60 \\
500   & 75.36 & 74.97 & 68.04 & 71.03 & 70.59 & 63.21 & 66.71 & 72.38 & 58.12 & 72.81 & 64.06 & 52.26 & 61.15 & 73.09 \\
1000  & 76.20 & 76.24 & 68.73 & 71.73 & 71.41 & 65.01 & 67.04 & 72.35 & 59.19 & 73.47 & 64.75 & 52.47 & 62.38 & 73.21\\
\bottomrule
\end{tabular}%
}

\resizebox{\textwidth}{!}{%
\begin{tabular}{lrrrrrrrrrr}
\toprule
XQ\textsc{u}AD & zh    & vi    & tr    & th    & ru    & hi    & es    & el    & de    & ar    \\
\midrule
0     & 48.14 & 49.02 & 36.90 & 27.84 & 51.86 & 42.47 & 54.48 & 42.90 & 56.22 & 46.40 \\
2     & 48.93 & 50.50 & 40.87 & 39.43 & 51.07 & 44.19 & 56.14 & 46.46 & 56.66 & 46.99 \\
4     & 49.72 & 51.38 & 40.22 & 41.24 & 51.33 & 45.90 & 56.62 & 47.25 & 56.38 & 46.57 \\
6     & 50.81 & 50.81 & 41.59 & 44.04 & 51.20 & 46.81 & 57.14 & 47.16 & 56.40 & 47.45 \\
8     & 51.53 & 51.29 & 41.99 & 45.28 & 51.29 & 47.10 & 57.45 & 47.95 & 57.07 & 48.21 \\
10    & 50.87 & 51.57 & 42.55 & 46.05 & 52.05 & 48.06 & 57.03 & 48.60 & 57.29 & 47.82\\
\bottomrule
\end{tabular}%
}

\caption{Final results with a different number of target-language data instances $k$, per language. Random sampling is used.}
\end{table*}

\clearpage

\section{Score gain trends}
\label{sec:app_gains}
\begin{figure*}[ht!]
    \begin{subfigure}{\textwidth}
        \centering
        \includegraphics[width=0.49\textwidth]{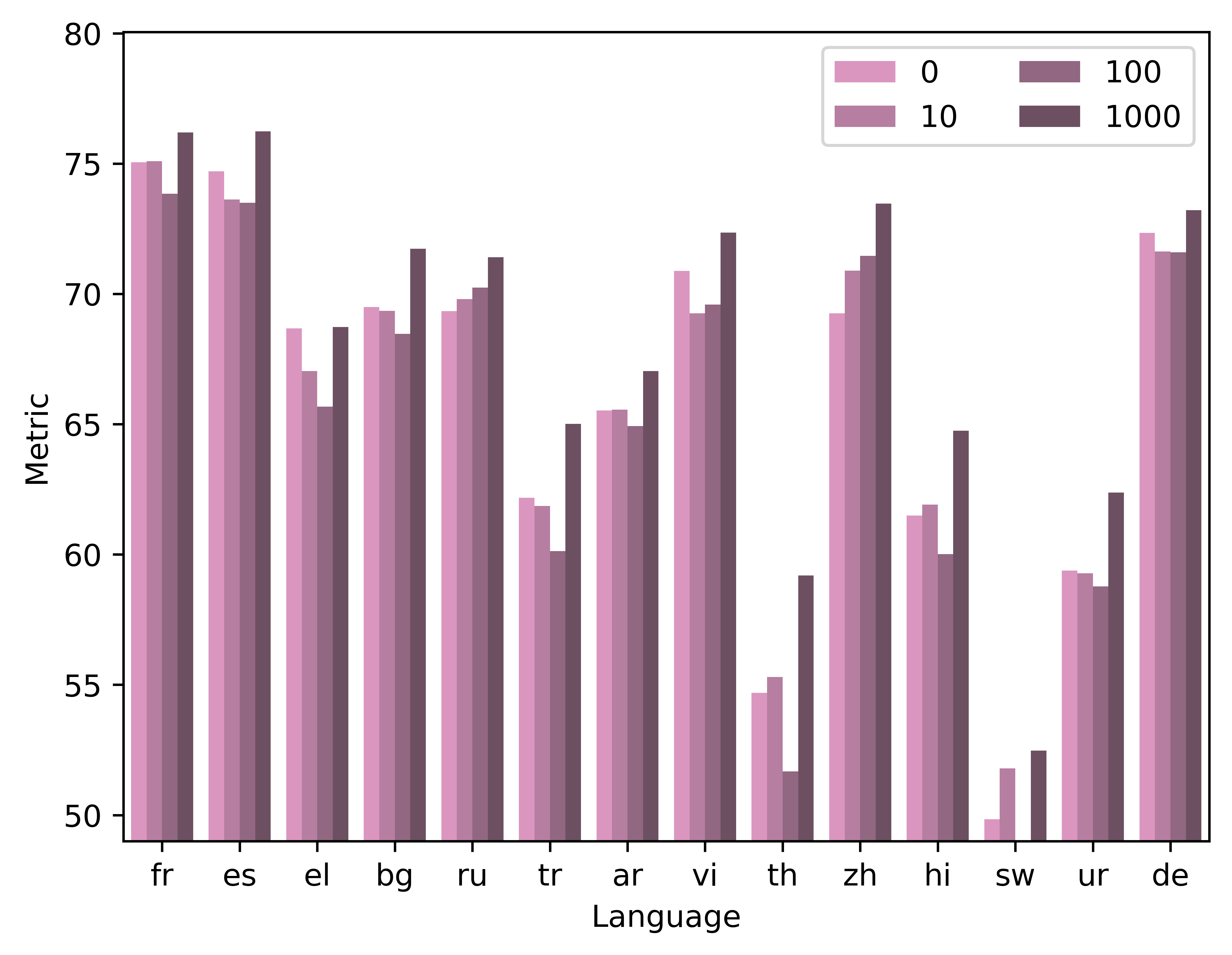}
        \caption{XNLI}
    \end{subfigure}
    \begin{subfigure}{0.49\textwidth}
        \includegraphics[width=\textwidth]{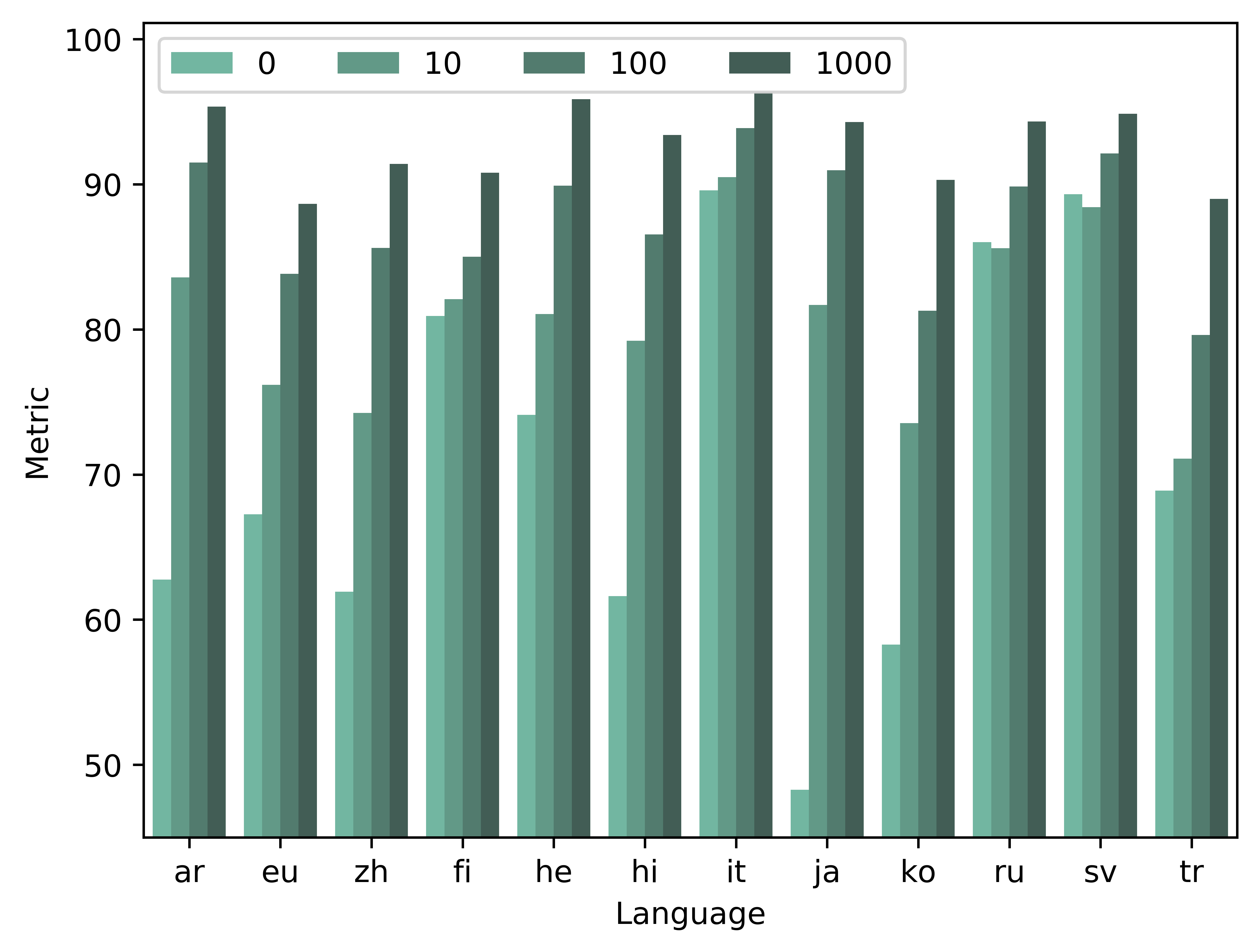}
        \caption{POS}
    \end{subfigure}
    \begin{subfigure}{0.49\textwidth}
        \includegraphics[width=\textwidth]{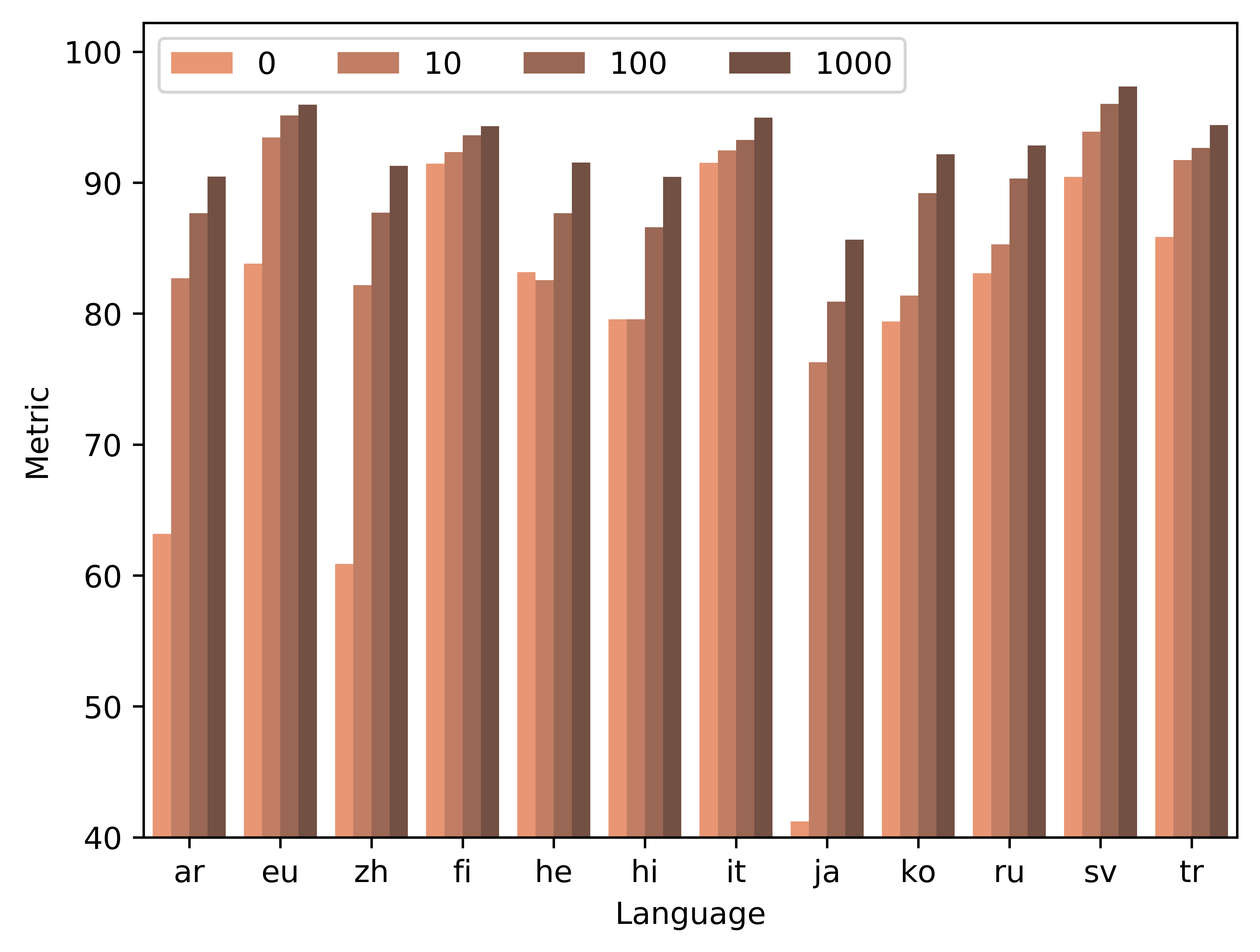}
        \caption{NER}
    \end{subfigure}
    \caption{Few-shot results for each language with varying $k$ for the remainder of tasks. All tasks report accuracy.}
    \label{fig:bar_all2}
\end{figure*}{}

\end{document}